# Point Spread Function Estimation of Defocus

Renzhi He, *Student Member, IEEE*, Yan Zhuang, Boya Fu, Fei Liu

*Abstract*—This Point spread function (PSF) plays a crucial role in many computational imaging applications, such as shape from focus/defocus, depth estimation, and fluorescence microscopy. However, the mathematical model of the defocus process is still unclear. In this work, we develop an alternative method to estimate the precise mathematical model of the point spread function to describe the defocus process. We first derive the mathematical algorithm for the PSF which is used to generate the simulated focused images for different focus depth. Then we compute the loss function of the similarity between the simulated focused images and real focused images where we design a novel and efficient metric based on the defocus histogram to evaluate the difference between the focused images. After we solve the minimum value of the loss function, it means we find the optimal parameters for the PSF. We also construct a hardware system consisting of a focusing system and a structured light system to acquire the all-in-focus image, the focused image with corresponding focus depth, and the depth map in the same view. The three types of images, as a dataset, are used to obtain the precise PSF. Our experiments on standard planes and actual objects show that the proposed algorithm can accurately describe the defocus process. The accuracy of our algorithm is further proved by evaluating the difference among the actual focused images, the focused image generated by our algorithm, the focused image generated by others. The results show that the loss of our algorithm is 40% less than others on average.

## I. INTRODUCTION

THERE is much research among the all-in-focus, focused, and depth images. Getting a clear image from a blurred image is called image deconvolution [1], [2], [3], [4], [5]; getting a depth image from a clear image or a focused image is called depth estimation [6], [7], [8]; the study from a series of focused images to a depth map is called shape from focus/defocus (SFF/ SFDF) [9], [10]. However, the relationship among these three is hard to describe intuitively.

From the perspective of computational photography, clear

This paragraph of the first footnote will contain the date on which you submitted your paper for review, which is populated by IEEE. This work is supported in part by National Natural Science Foundation of China (No.T2222018) and Innovation group science fund of Chongqing Natural Science Foundation (cstc2019jcyj-cxttX0003). The corresponding author is Fei Liu.

Renzhi He, Boya Fu, Fei Liu are with the School of Mechanical and Vehicle Engineering, Chongqing University and State Key Laboratory of Mechanical Transmission, E-mail: fei_liu@cqu.edu.cn; cubhe@foxmail com; boya_f@163.com

Yan Zhuang is with Chongqing University-University of Cincinnati Joint Co-op Institute E-mail: 20186105@cqu.edu.cn

image, also known as all-in-focus image, can be regarded as a projection of the original scene. A focused image is derived from the original scene. The defocus scale is related to the depth of the object and focus depth. The defocus process can be described by point spread function (PSF). Thus, the PSF can effectively unify these three, which generate the focused image using a depth map and an all-in-focus image.

Although there has been lots of research on the PSF [1], [11], [12], [13], due to some optical parameters cannot be measured directly, such as the f-number, the physical size of pixels, etc. and some data being hard to acquire, such as the all-in-focus image, focused image with corresponding focus depth, and depth map in the same view, the precise PSF is hard to get. To complement and advance previous work [10], [11], [13], we aim to obtain a precise PSF model of the camera.

There are two approaches to get the point spread function, physically-based and computational-based. The physically-based methods usually acquire the images of the point-like light or spots [12]. Due to the low SNR of images, the PSF is far from accurate. However, the physically-based method can be improved by optimizing the hardware, such as the aperture or CCD/CMOS sensors [14], [15], [25]. As for the computational methods, the image data, all-in-focus image, focused image, depth map are inaccurate and hard to measure accurately [11], which leads to the inaccurate of the PSF. These methods are usually combined with statistical methods [14], [15], [16] to overcome the inaccuracy of datasets.

According to the existing research, it is difficult to obtain the all-in-focus image, focused image with corresponding focus depth, and depth image of the same viewpoint, so it is hard to quantifiably evaluate the accuracy of the algorithm.

We proposed a novel approach to acquire PSF based on precision measuring instruments and accurate mathematical models, Fig.1. Firstly, for the data that were previously difficult to obtain in the model, we designed an experimental equipment, which consists of a focusing system and a structured light system. This equipment can obtain the all-in-focus image, focused image with corresponding focus depth, and depth map in the same view. The focusing system obtains the focused image with corresponding focus depth. The structured light system consists of a projector and a camera, which obtains the all-in-focus image and the depth map.

After obtaining the image dataset, we use it to acquire the PSF. The PSF is modeled with the Gaussian function [23], [26], [30]. Then we extract two parameters that need to be solved.

One is the optical parameter ($A$) which is considered as a composite of camera's f-number ($N$), pixel size ($\rho$), output scale ($s$), and scaling factor of circle of confusion ($\omega$). Detail discussion of $A$ is in Sect. 3. The other is the mechanical parameter ($e$) which is the deviation of the focus depth and will be discussed in Sect. 4.

We solve these two paraments by minimizing the loss function. Besides the previously used luminance loss, such as L1 and L2 loss, and defocus loss based on Laplacian or Sobel loss, we design a new loss function called the defocus histogram loss, which is more sensitive to variations in defocus. To boost the computational performance, the algorithm is implemented using CUDA tools on GPUs.

We conduct experiments both on standard planes and actual objects. To further validate our algorithm, we take the quantitative analysis for the actual focused image, the focused image generated by our algorithm, and the focused image generated by others [23].

To summarize, our contributions in this work are listed as follows:

- We set up an image dataset consisting of 30 sets of data in which 6 sets are standard planes and 24 sets are actual objects. Each set contains the all-in-focus image, focused images with corresponding focus depth, and depth map from the same view without changing the camera imaging model. We make our dataset publicly available and keep it updated.
- We derive the mathematical model of the camera's PSF. With the image dataset, the precise PSF is obtained.
- We designed a new metric (defocus histogram loss) to quantitatively analyze the focused images. It is more sensitive to the variations in defocus and get better converge than the general luminance loss and defocused loss.

The rest of the paper is organized as follows. In Section II, we discuss the related work. Section Ⅲ gives an analysis of the camera's optical model and derives the PSF function. Dataset acquiring and the hardware system are introduced in Section IV. The details of the algorithm implementation will be discussed in Section V. Experiments on the standard planes and objects are given in Section VI to show the algorithm's performance. Finally, we summarize and discuss the future research in Section VII.

## II. RELATED WORK

**Physically-based methods.** Subbarao *et al.* [13] proposed a computational model for image sensing in a typical CCD camera system. However, many paraments in the model are unknown, such as the distance between the exit pupil and the image sensor, etc. They did not perform a quantitative analysis of the model. Some methods usually acquire the images of the point-like light or spots [12]. Due to the low SNR of images, the PSF is far from accurate. Some methods optimize the hardware to improve the result. [14] analyze the PSF based on dual-pixel sensor [15] use the coded exposure PSF. Instead of changing the imaging hardware, we aim to find a more general way to estimate the PSF of a camera.

**Shape from focus/defocus (SFF/SFDF).** In the shape from focus methods, the SFF retrieves the best sequence based on the degree of defocus, and the PSF is used to assist this retrieval process as a fitting function [20], [21], [22]. In the SFDF method, the result is more sensitive to the PSF. Due to the inaccuracy PSF and noise, it is difficult to get precise depth with SFF/SFDF.

**Depth estimation from defocus cue.** In neural network-based methods, the depth of the image is estimated using different levels of defocus scale of the image [18], [23], [24], [25], [26]. Some methods also incorporate physical models to make the results more accurate [27].

These algorithms inevitably use the all-in-focus image, focused image, and depth map. However, there is still not a precise PSF of the camera or a precise dataset of the all-in-focus image, focused image with focus depth, and depth image of the same view. Anmei Zhang *et al.* [31] proposed a "Defocus-Depth Dataset", which is labeled by boxes rather in pixel level. Srinivansan *et al.* [32] presented a lightfield dataset. However, it does not contain the focus depth and depth map. Many works render the focused image from the RGB-D dataset, such as NYU Depth Dataset V2 (NYU-v2) [40], where the depth maps are acquired by Microsoft Kinect or laser scanner. In the render process, a precise PSF is required. In our work, we provide the high-resolution depth map and a more precise PSF.

**Comparison with our approach.**
We build an experimental equipment consisting of a structured light system and a focusing system to obtain a precise dataset of all-in-focus image, focused image with corresponding focus depth, and depth image in the same view. The structured light system allows us to obtain highly accurate depth images with 0.1mm resolution, which is more accurate than the NYU-v2 [40] and Make3D [41]. The focusing system can obtain high-resolution focused images of 2096*2000(pixel) and focus depth with 0.003mm resolution.

A precise PSF of a single-lens camera is built and then solved on our dataset. We solve the composited optical parameter of camera's f-number ($N$), pixel-size ($\rho$), output scale ($s$), and scaling factor of the circle of confusion ($\omega$) in the PSF, rather than use the value given on the hardware manual. Moreover, we do not need to customize the aperture or CCD sensor of the camera, which means our method is more general. Many works, such as SFF/SFDF, image deconvolution, depth estimation, are sensitive to the PSF. However, they use the given values to generate the PSF, which is not that accurate in Sect.6.4. We hope our work, a precise PSF and the dataset, will facilitate these works.

Moreover, to improve previous loss functions, we design a new metric based on the defocus histogram to quantitatively evaluate the focused images, which is more sensitive to the variations in defocus and get better converge than luminance loss and defocused loss, and more efficient.

## III. POINT SPREAD FUNCTION

This section introduces the defocus process, where the focused image can be seen as a superposition of the circle of confusion (CoC). Firstly, we introduce the optical principle of CoCs. Then we derive the mathematical model of CoCs, which is known as point spread function based on the Gaussian function. Finally, we derive the mathematical model of the focused image.

### A. Circle of Confusion

Considering the symmetrical lens as illustrated in Fig. 2, the focal length $F$, the distance from the optical lens to sensor plane $v$ and the focus depth $D_f$ satisfy the following equation:

$$\frac{1}{D_f} + \frac{1}{v} = \frac{1}{F} . \tag{1}$$

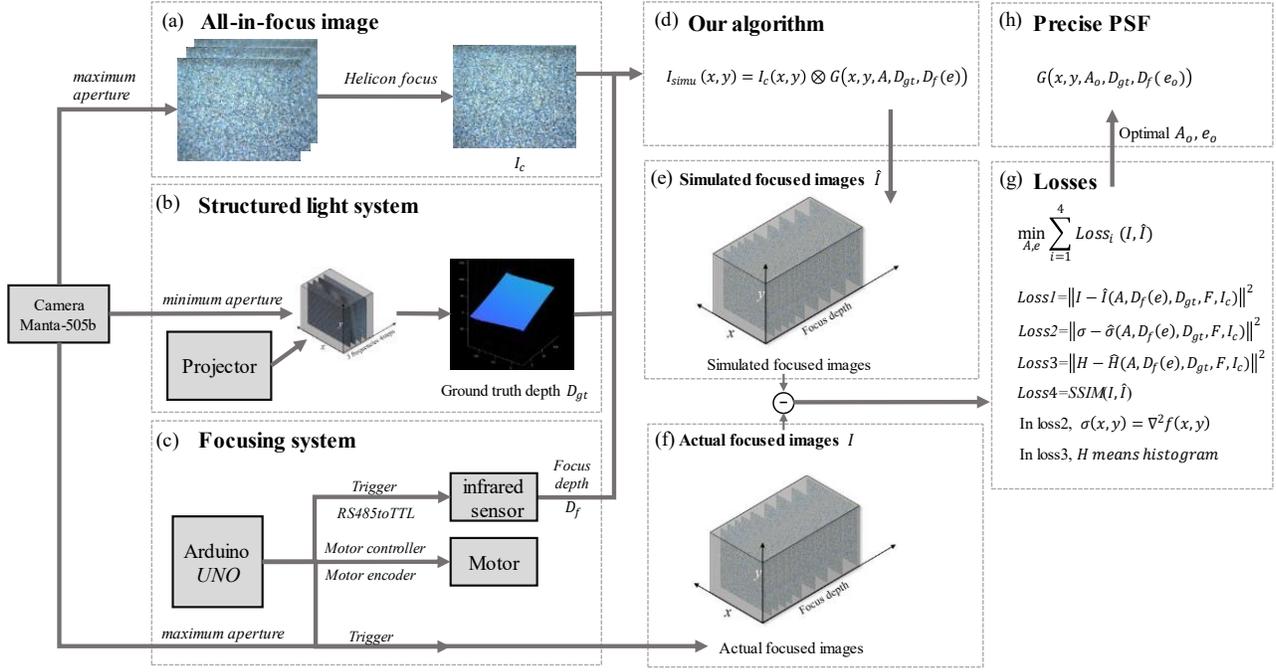

**Fig. 1.** Implementation of the proposed PSF estimation system. (a) All-in-focus image, which can be synthetic from the focused images or get from the camera with minimum aperture. (b) Structured light system: The structured light system consists of a projector and a camera. (c) The focusing system: The focusing system consists of the same camera in (b) with maximum aperture, a motor to control the zoom, and an infrared sensor to measure the focus depth. (d) The point-spread-function implementation (more details in Sect. 3). (e) Simulated focused images generated by (d). (e) Actual focused images captured by the focusing system in (c). (g) Loss function in Sect. 5.2. (h) Precise PSF model

If the positions of the sensing plane and focus plane do not coincide, the light from the object point projects a Circle of Confusion on the sensing plane.

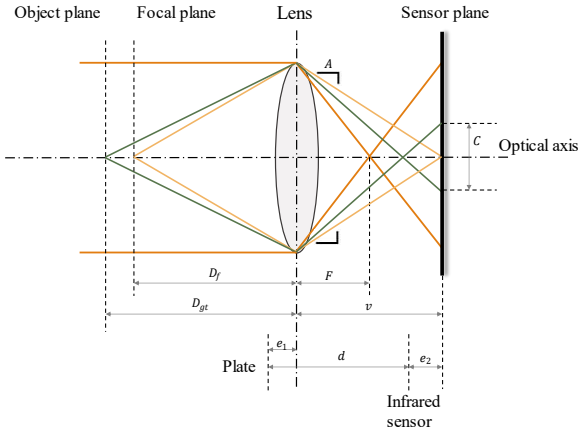

Fig. 2. A simple model of focus with geometric optics. The orange line represents the incident light as parallel light when the light will be focused on the lens focus. The yellow line represents the focal point when the light passing through the lens will form a point on the sensor plane. The green line represents the point that is not in focus, the light passing through the lens will form a diffuse circle in the focus plane, where the diameter of the CoC is represented by $c$.

As seen from Fig.2, the diameter of the CoC can be written as:

$$C_{mm} = a \frac{|D_{gt} - D_f|}{D_{gt}} \frac{F}{D_f - F} \tag{2}$$

where $D_{gt}$ is the distance between an object and the lens, and $a = F/N$ where $N$ is the f-number of the camera. While CoC is usually measured in millimeters ($C_{mm}$), we transform its size to pixels by considering a camera pixel-size $\rho$, and a camera output scale $s$. The CoC size in pixels $C_{pix}$ can be rewritten as:

$$C_{pix} = \frac{C_{mm}}{\rho \cdot s} = \frac{a}{\rho \cdot s} \frac{|D_{gt} - D_f|}{D_{gt}} \frac{F}{D_f - F} \tag{3}$$

*B. Point Spread Function*

In fact, due to the influences of optical interference and diffraction, CoC has a bright center and blurring edges rather than a uniformly bright disk. The response of an imaging system to a point source or object is represented by a PSF, which can be approximated by the following Gaussian function, [23], [26], [30]:

$$G(x, y, r) = \frac{1}{2\pi r^2} \exp\left(-\frac{x^2 + y^2}{2r^2}\right). \tag{4}$$

In this work, we consider the defocused model to be a disc-shaped PSF. Following [23], [27], [30], we assume that $r$, the only parameter in the Gaussian function, is related to the diameter of the CoC, $C_{pix}$. Instead of replacing $r$ with $C_{pix}$, we consider that $r$ can be expressed as $r = \omega \cdot C_{pix}$. Combining (4) with the diameter of CoC in (3):

$$G\big(x, y, A, D_{gt}(x,y), D_f\big) =$$

$$\frac{1}{2\pi\left[\frac{a}{\rho \cdot s}\frac{|D_{gt} - D_f|}{D_{gt}}\frac{F}{D_f - F}\right]^2}\exp\left[-\frac{x^2 + y^2}{2\left[\frac{a}{\rho \cdot s}\frac{|D_{gt} - D_f|}{D_{gt}}\frac{F}{D_f - F}\right]^2}\right]. \quad (5)$$

For the $a, \rho, s$, many works[23], [27], [30] use the parameters given in the hardware manual and some empirically bring $C_{pix}$ to $r$. In our algorithms, we consider these parameters as an optical parameter $A$, and treat it as an unknown parameter

$$A \coloneqq \frac{a \cdot \omega}{\rho \cdot s}. \quad (6)$$

### B. Point Spread Function

When all points in the real scene are exactly at the focus depth of the lens, an all-in-focus image is generated where each point is independent of the other. But when the ground-truth depth does not coincide with the focus depth, then a CoC is generated and the point affects the points around it. The focused image can be regarded as the superposition of CoCs.

We first compute the CoC for each point and construct a kernel for each point that every item in the kernel represents the effect of the corresponding point's CoC on the target point. Then the point in the focused image can be modeled as the convolution of the all-in-focus image with the kernel. The focused image can be modeled as:

$$I_b(x, y) = I_c(x, y) \otimes G\big(x, y, A, D_{gt}, D_f\big) \quad (7)$$

$$I_b(x,y) = \sum_{i=x-\frac{k-1}{2}}^{x+\frac{k-1}{2}}\sum_{j=y-\frac{k-1}{2}}^{y+\frac{k-1}{2}} I_c(i,j) \cdot G\big(x - i, y - j, A, D_{gt}(i,j), D_f\big) \quad (8)$$

where $k$ is the kernel size, $I_b(x,y)$, $I_c(x,y)$ are focused image and all-in-focus image. $I_b, I_c, D_f, D_{gt}$ will be measured in Sect. IV. $A$ will be solved by the algorithm in section V, kernel size $k$ is odd and discussed in Sect VI.

## IV. DATASET

We design a hardware system consisting of a structured light system and a focusing system to obtain all-in-focus images, focused images with corresponding focus depth, and depth images in the same view.

### A. Hardware

The structured light system consists of an Acer H6542BD projector, a Qiyun 5014-8MP lens, and a Manta G-505 camera. The focusing system, as shown in Fig. 3, shares the same camera and lens as the structured light system. An ECOTTER LSD-50-RS-485 infrared sensor with a resolution of $3\mu m$ is used to measure the focus depth. A stepper motor is used to control the zoom. Finally, an Arduino is used as the main controller to trigger the camera and the infrared sensor, as well as control the stepper. A computer is connected to the Arduino to acquire and analyze these data.

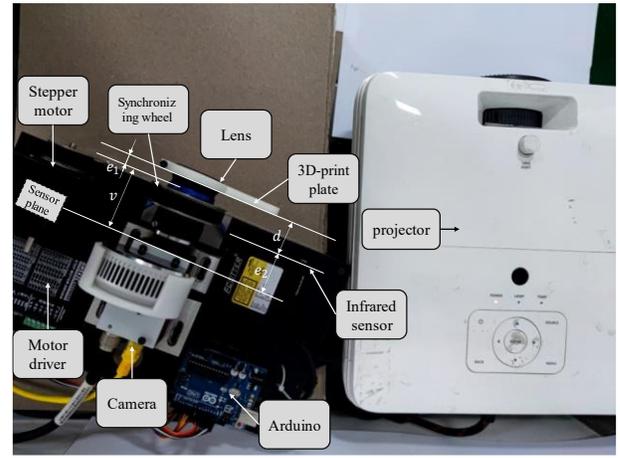

Fig. 3. The hardware system for our model. The structured system consists of a projector and a camera. The focusing system mainly consists of an Arduino, an infrared sensor, a camera, and a stepper motor. The infrared sensor measures the distance to the plate, as noted as $d$. The distance between the lens and the CCD sensor is noted as $v$.

### B. Data Acquisition

Our dataset consists of *all-in-focus image, focused image, object depth map, and focus depth*. It is challenging to get the focused image and depth map in the same view.

**Focused image:** The focused images are captured by the focusing system with the maximum aperture $\#F=1.4$, as shown in Fig. 4. With the maximum aperture, we can get focused images with a small depth of field images. The system obtains a focused image and its corresponding focus depth. We add a 3D-printed flat plate at the front of the lens, Fig. 3. The focus depth is then obtained using a fixed infrared sensor by measuring the distance between the infrared sensor and the flat plate. A custom synchronous wheel, whose movement is controlled by a motor, is mounted on the zoom ring.

**All-in-focus image:** There are two approaches to obtain all-in-focus images. The first approach is to use the camera with a minimum aperture $\#F=16$ to take a photo. With the minimum aperture, we can get images with a large depth of field images. Combined with the structured light method, the large depth of field images can be used for solving the depth map.

The second approach is to synthesize an all-in-focus image from focused images by using the Helicon Focus software or our algorithm, which is based on the SFF method of finding clarity [22]. Based on our experience, the all-in-focus images built by the second approach are closer to the original scene.

**Depth map:** The structured light system obtains the object's depth map. As shown in Fig. 5, four phase-shifted images of each of the three frequencies are projected to the object and then the camera capture the stripes. We solve the depth based on fringe projection profilometry [33]. However, the depth of the structured light is based on the world coordinate system, and the depth needs to be converted to the camera's coordinate through the extrinsic matrix. It is worth noting that the X-Y plane and the origin of the depth map need to be consistent with the image.

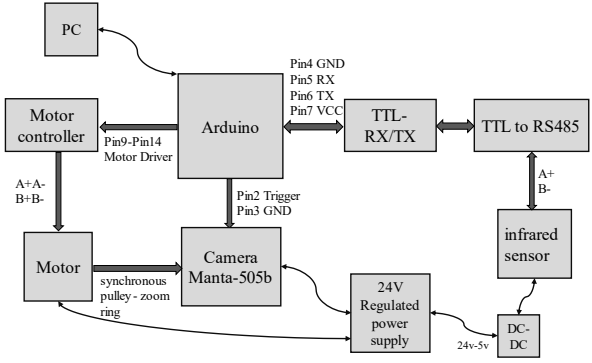

Fig. 4. The electrical structure of the focusing system. The main control is Arduino, which is connected to a motor driver, an infrared sensor and the camera trigger. A 24V DC regulated power supply with a step-down module is used to power the whole system.

where $R_{3\times3}$ and $T_{3\times1}$ are the rotation and translation matrices from the world coordinate system to the camera coordinate system, which can be obtained by Zhengyou Zhang's calibration [35]. $P_{3\times1}^{word}$ is the object position in the world coordinate, $P_{3\times1}^{cam}$ is the object position in the camera coordinate.

**Focus Depth**: The infrared sensor measures the distance from the plate to the infrared sensor. It does not directly measure the focus depth. There is a particular gap $e$ between the measured distance $d$ and the distance from the optical lens to CCD sensor plane $v$, as shown in Fig.2 and Fig. 3. $e$ consists of two parts. The distance between the infrared sensor and the CCD sensor is $e_1$, the distance between the 3D-printed plate and the lens is $e_2$ .

$$e = e_2 - e_1 \qquad (10)$$

The distance from the optical lens to CCD sensor plane $v$:

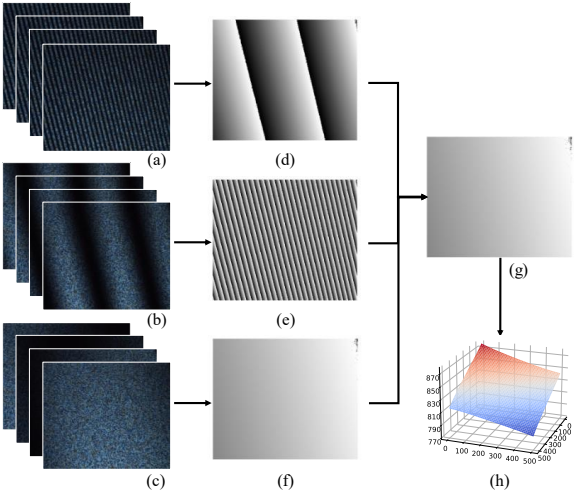

Fig. 5. The process of fringe projection profilometry. (a) ~ (c) are 12 images captured by the camera, containing 4 phase-shifted images for each frequency. Then a relative phase is calculated for each frequency, as shown in (d) ~ (f). The absolute phase (g) of the three frequencies is then derived from the relative phase. Finally, the absolute phase is calculated, and the object depth is solved according to the triangular principle of light (the origin of the coordinate system is established on the calibration plate when calibrating the external reference).

$$v = d + e \qquad (11)$$

Seen from Fig.2 and Fig.3, $e_1$ and $e_2$ are constants. Therefore, one way to roughly estimate $e$ is to solve $v$ by Zhengyou Zhang's calibration method. Another more accurate algorithm is to use the optimization algorithm proposed in Sect. 5 to find the gap $e$. In any case, when we get $e$ and combined with (1) the depth of focus, $D_f$ can be obtained by the following equation:

$$D_f = \frac{f(e+d)}{e+d-F} \qquad (12)$$

## V. Model Architecture

In this section, we present the pipeline of our model and discuss the loss function, as shown in Fig. 1. We aim to find a precise PSF for the camera using the optimization method, which minimizes the loss function to solve the optical paraments $A$ in Sect. 3 and the mechanical parameter $e$ in Sect. 4. Our model consists of 3 parts, focused image generation algorithm, loss function, and an iterative method.

### A. Algorithm

The focused images are generated following (8). Instead of generating a Gaussian function-like kernel, we first compute the CoC for every point and construct a kernel where each item in the kernel is related to the corresponding CoC. Since we need to generate a kernel for each point, compute each item in the kernel, and compute the convolution of every point with the kernel, the PyTorch and the Cuda tools are used to compute these in parallel by GPUs.

There are two parameters that need to be solved, the optical parameter $A$ mentioned in Sect.3 and the mechanical parameter $e$ mentioned in Sect. 4. We rewrite the focused image generation function $\mathcal{F}$ to meet the optimization algorithm, as

$$I_b = \mathcal{F}\big[A, D_f(e), D_{gt}, F, I_c\big] \qquad (13)$$

$$I_b(x,y) = I_c(x,y) \otimes G\big[x, y, A, D_f(e), D_{gt}(x,y), F, I_c\big] \qquad (14)$$

where the $I_c$ is the all-in-focus image, $I_b$ is the focused image, $D_f$ is focus depth in (12).

### B. Losses

Losses play a critical role in measuring the difference between two focused images in this model. The choice of losses and ratio between them directly determine whether the algorithm can find the optimal solution of the system or not. Many deconvolution algorithms adapt a loss function which is the weighted sum of MSE/L2 loss, SSIM loss [36], and defocuses/Sobel loss [37], [38] to estimate the difference between blur images and clear images. However, after our experiments, we found that using L2 loss and defocus loss is insufficient to distinguish the difference between two focused images. Thus, we designed a new loss function called defocus histogram loss to overcome the shortcomings of the previous losses. Moreover, we added SSIM to eliminate the effect of slight zoom on the image. We compute the loss between the actual focused images and the simulated focused images which are generated using different optical and mechanical parameters, as shown in Fig.

6.

**Luminance loss (loss1).** The Luminance loss, also called L2 loss, can describe the difference between the two images, as (15).

$$Loss1 = \left\| I - \hat{I}\left(A, D_{f_i}(e), D_{gt}, F, I_c\right) \right\|^2 \tag{15}$$

where the $\hat{I}$ is the simulated focused image, $I$ is the actual focused image.

However, we compare the two focused images in our model, which are only slightly transformed from the all-in-focus images. The luminance loss of any two focused images is small, so it is insufficient to evaluate the difference. Thus, we introduce two defocus-related losses

**Defocus loss (loss2).** Since we evaluate the differences between two images with different levels of defocus, it is necessary to evaluate the levels of defocus of each point. As the levels of defocus are related to sharpness, we use the Laplace kernel to compute the sharpness at each point and compute the L2 loss of the two sharpness maps.

$$Loss2 = \left\| \sigma - \hat{\sigma}\left(A, D_{f_i}(e), D_{gt}, F, I_c\right) \right\|^2 \tag{16}$$

where the $\sigma$ denotes the Laplacian of the image

**Defocus histogram loss (loss3).** On the basis of luminance loss and defocus loss, we find that our model still does not get a good convergence, as shown in Fig. 6(a) and Fig. 6(b). Because the defocus loss calculates the average of the sharpness, it cannot represent the sharpness distribution. Moreover, when the optical parameter $A$ is large, the transition from focused to defocused is very drastic. In order to find the optimal value of the optical parameter $A$, the loss function needs to be sensitive to the distribution of the sharpness. We designed a histogram to record the number of pixels in each sharpness scale. We calculate the mean square error between two images' sharpness histograms and name it the defocus histogram loss. As shown in Fig. 6(c), when $A$ increases, the loss starts to increase and then decreases. The loss function get convergence at one point, which means the defocus histogram loss is more sensitive to the distribution of the sharpness than loss1 and loss2. Loss3 can be written as:

$$Loss3 = \left\| H(\sigma) - H\left[\hat{\sigma}\left(A, D_{f_i}(e), D_{gt}, F, I_c\right)\right] \right\|^2 \tag{17}$$

where $H$ means the histogram function.

**SSIM loss (loss4).** The structural similarity loss, a statistically based loss, can be used to evaluate the clarity of images [36]. Because our algorithm generates the focused image from the all-in-focus image, there is a slight difference between the simulated and the actual focused images in the field of view. The image will be slightly deflated when changing the camera's focus depth. In order to eliminate the effect, we introduce the SSIM loss to evaluate the difference between the images.

$$Loss4 = \left(1 - SSIM(I, \hat{I})\right)/2N \tag{18}$$

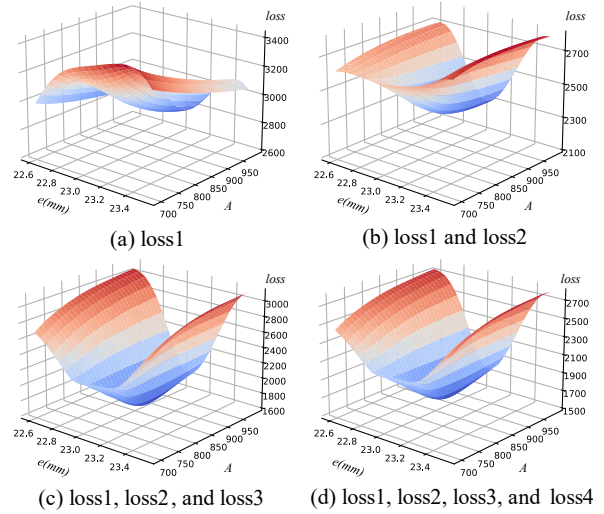

Fig. 6. The results of using different losses. (a) Only loss1 is used. (b) Loss1 and loss2 are used. (c) Loss1, loss2 and loss3 are used. (d) All four losses are used.

where SSIM is the structured similarity loss function[36], N is the number of pixels.

Fig. 6. shows the effect of each loss. Using loss1 and loss 2 cannot get convergence. When $A$ is large, these losses cannot evaluate the difference between the simulated focused image and the actual focused image. After introducing loss3, as shown in Fig. 6(c), the loss function gets convergence. In Fig. 6(d), the SSIM loss is added to the loss function. We use the four losses to evaluate the difference between focused images: the luminance loss, the defocus loss, the defocus histogram loss, and the structural similarity loss. The loss function can be written as:

$$Loss = \lambda_1 Loss1 + \lambda_2 Loss2 + \lambda_3 Loss3 + \lambda_4 Loss4. \tag{19}$$

Empirically, we set the $\lambda_1$, $\lambda_2$, $\lambda_3$, $\lambda_4$ as 50 000, 10000, 40000, 5000, respectively.

### C. Optimization method

After obtaining the simulated focused images for each focus depth, we compare these images with the actual focused images captured by the focusing system. We refer to our losses to represent the difference between two focused images and use (19) to evaluate the difference between them. When we find the minimum value of the loss, it means the focused images generated by our algorithm are close to the actual focused images. In other words, we solve:

$$s(A,e) = arg\ \min_{A,e} \left\{ \sum_{i=1}^{M} loss\left[I, \hat{I}\left(A, D_{f_i}(e), I_c\right)\right]/M \right\} \quad (20)$$

where $M$ represent the number of focused images. $A$ is the optical parameter defined in (6) and the mechanical parameter $e$ is defined in (10). Then we use the two-dimensional exhaustive search to solve the 2 parameters in the model.

TABLE 1
ALGORITHMS

| Algorithm 1: Iterative search |
|---|
| **Input:** lower and upper mechanical parameter $e_{min}$, $e_{max}$, lower and upper optical parameter $A_{min}$, $A_{max}$, all-in-focus image $I$, focused depth $D_f$, total image number $M$, simulation function $F$, loss function $L$ |
| **Output:** optimal mechanical parameter $e_p$, optimal optical parameter $A_p$ |
| 1   Initialize minimal loss $l \equiv +\infty$; |
| 2   Initialize optimal mechanical parameter $e_p \equiv 0$; |
| 3   Initialize optimal optical parameter $A_p \equiv 0$; |
| 4   **for** *image identity* $D_f = 1, \ldots, M$ **do** |
| 5     **for** $A = A_{min}, \ldots, A_{max}$ **do** |
| 6       **for** $e = e_{min}, \ldots, e_{max}$ **do** |
| 7         Simulate focused image $I_s$: $I_s = \mathcal{F}(A, e, D_f, I)$; |
| 8         Load real focused image $I_r$; |
| 9         Calculate loss between $I_s$ and $I_r$: $l_{cur} = L(I_r, I_s)$; |
| 10         **if** $l_{cur} < l$ **then** |
| 11           Update minimal loss $l = l_{cur}$; |
| 12           Update optimal mechanical parameter and optical parameter $e_p = e$, $A_p = A$ |
| 13        **end** |
| 14     **end** |
| 15   **end** |

## VI. EXPERIMENTS

### A. Setup

We present a set of experiments both on the standard planes with synthetic patterns and actual objects. We use the method proposed in Sect. 4 to obtain the depth of these objects, all-in-focus images, focused images, and the focus depth of each focused image.

We implement the model using the open-source PyTorch framework and the Cuda tools with an Nvidia Titan RTX 24G.

**Kernel size.** After we derive the mathematical model of the CoC, we need to digitize the kernel. In the process, most of the algorithms use a fixed size of the kernel. However, this kernel ignores points when the CoC is large and includes extra points when the CoC is small. We use a dynamic kernel size that transforms with the size of the CoC. We set the dynamic factor to 4, which means that the size of the convolution kernel is numerically 4 times the radius of the circle of confusion. Fig. 8 shows the difference between the dynamic kernel size and the fixed kernel size. The minimum loss result of the dynamic convolution is 1657. It is smaller than that of the fixed convolution kernel (1921).

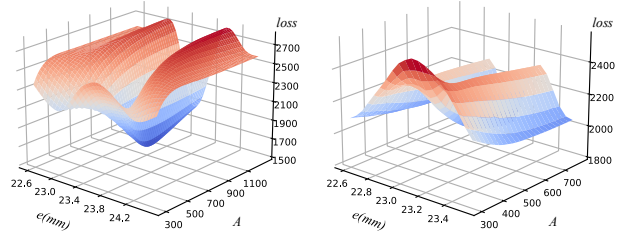

Fig. 7 (a) use the dynamic kernel size with minimum loss 1657; (b) use the fixed kernel size with minimum loss 1921 and does not converge.

### B. Tests on Standard Planes

We generate 5 simulated masks and paste them on the standard planes. The extensive textures allow the focusing system to obtain high-quality focused images. The standard planes allow the structured light system to obtain high-quality depth map.

However, the chromatic aberration is different for different wavelengths of light [24], which will affect the CoC's location of focused images. In order to reduce its effects, we generate the simulated masks using white and 5 colors which are taken evenly on the hue circle and distributed randomly in the mask. Table. 2 shows the colors we used to generate the mask. The 3 values of each item are the value of the red, green, blue channel. The results are shown in Fig. 9 and Table 3. However, the chromatic aberration is different for different wavelengths of light [24], which will affect the CoC's location of focused images. In order to reduce its effects, we generate the simulated masks using white and 5 colors which are taken evenly on the hue circle and distributed randomly in the mask. Table. 2 shows the colors we used to generate the mask. The 3 values of each item are the value of the red, green, blue channel. The results are shown in Fig. 9 and Table 3. For the experiments on the standard planes, due to the standard plane and the textures, the results for every set of data can get convergence in the same point. The shape around the minimum value is very sharp, seen the loss result of masks in Fig. 9, which means

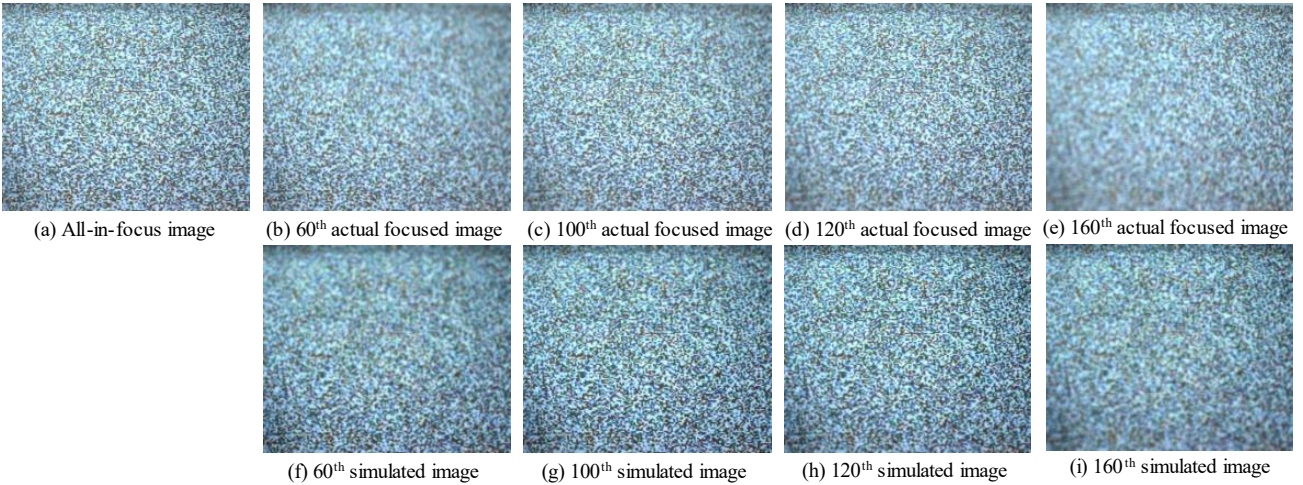

(a) All-in-focus image    (b) 60th actual focused image    (c) 100th actual focused image    (d) 120th actual focused image    (e) 160th actual focused image

(f) 60th simulated image    (g) 100th simulated image    (h) 120th simulated image    (i) 160th simulated image

Fig. 8. Illustration of the comparison of the actual focused image and the simulated focused image. (a) the all-in-focus image, (b)-(e) are the 60th 100th 120th 160th images in the actual focused image, respectively, and (h)-(i) are the 60th 100th 120th 160th images in the simulated focused image.

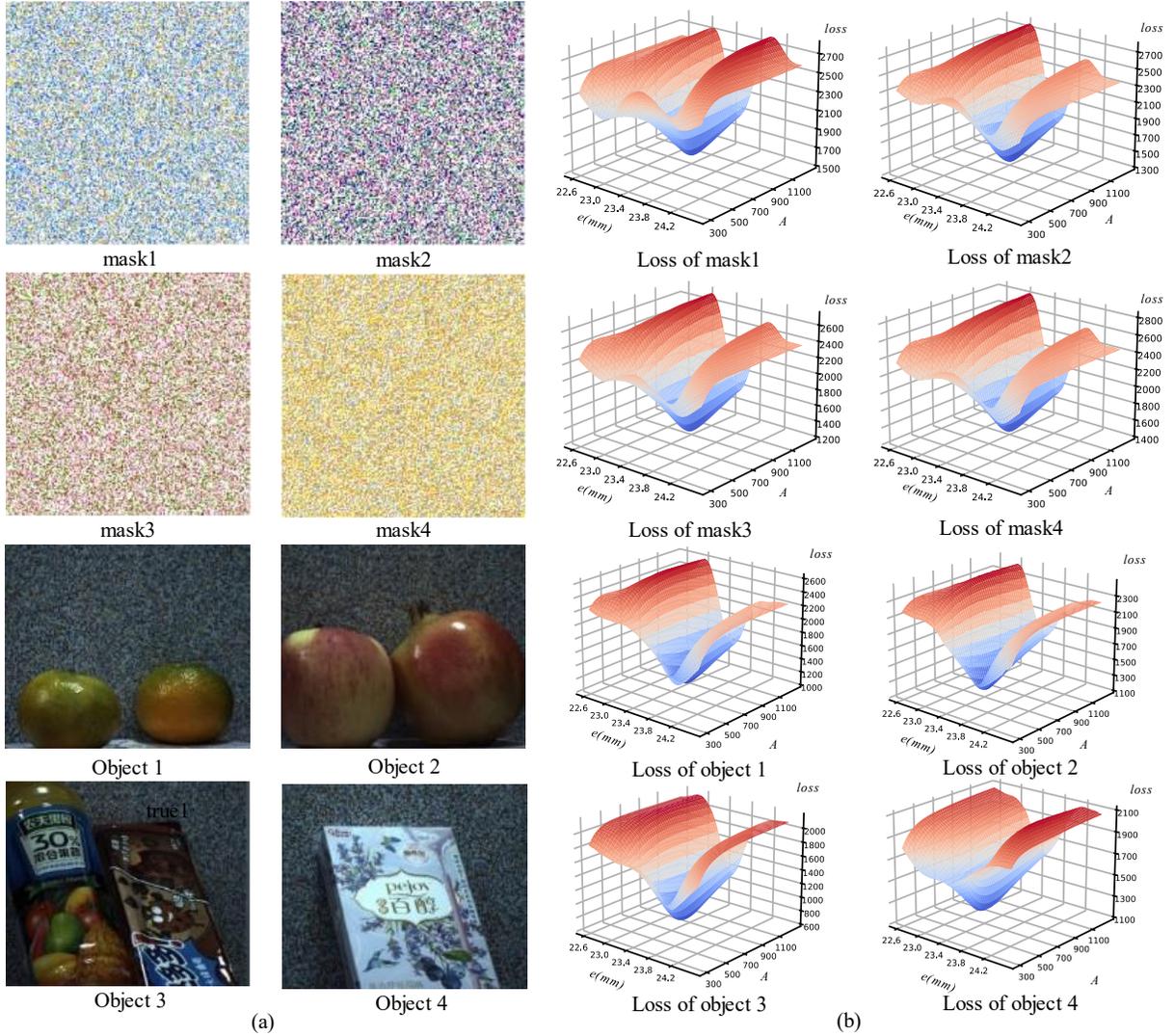

Fig. 9. The illustration of the loss results of the standard planes and actual object. (a) The simulated masks and the actual objects we used to evaluate our model. (b) The corresponding loss results of (a).

that the results are accurate and the simulated focused images are close to the actual focused images, as shown in

TABLE 2
THE COLORS OF STANDARD PLANES

|  | Color1 | Color2 | Color3 | Color4 | Color5 |
|---|---|---|---|---|---|
| Mask1 | 255, 255, 0 | 255, 80, 80 | 255, 0, 255 | 54, 97, 143 | 51, 204, 204 |
| Maks2 | 184, 124, 76 | 126, 0, 0 | 153, 51, 255 | 217, 243, 255 | 51, 204, 51 |
| Maks3 | 226, 182, 89 | 255, 229, 222 | 204, 102, 255 | 132, 174, 225 | 22, 73, 117 |
| Maks4 | 255, 219, 118 | 255, 117, 194 | 255, 93, 29 | 0, 177, 249 | 134, 117, 131 |
| Maks5 | 255, 247, 208 | 255, 179, 170 | 255, 112, 148 | 0, 106, 247 | 0, 255, 80 |

Fig. 11(a) and Fig. 11(b). At the minimum point, the optical parameter $A$ is 800, and the mechanical parameter $e$ is 23.6mm. We generate some focused images with the optimal parameters, Fig. 8. It shows that the actual images and the simulated images are similar, especially the defocus areas are very close.

### C. Tests on Actual Objects

For the actual object experiments, we choose some fruits and boxes. The loss results are shown in Fig. 10 and Table 3. The loss function can find the minimum values.

Seen from Tab. 3, the optical parameters $A$ are slightly different among different objects, and the mechanical parameter $e$ is 23.6 mm of all object data. For optical parameters $A$, as we discussed in Sect. 6.2, the color has an effect on $A$. For the data of masks, the color in the mask is evenly taken on the hue circle and distributed randomly. Thus, each color has the same effect on the optical parameter. The optical parameters of the masks are the same. But for the actual object, the object image tends to be dominated by a certain color, which causes the optical parameter a little different on different object.

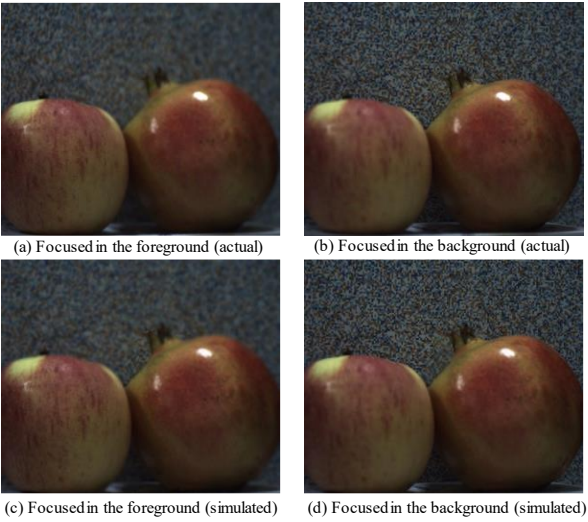

(a) Focused in the foreground (actual)   (b) Focused in the background (actual)

(c) Focused in the foreground (simulated)   (d) Focused in the background (simulated)

Fig. 10. The actual focused images are compared with the simulated images. (a) (c) shows the actual picture with the focus in the foreground vs. the simulated picture. (b) (d) shows the actual with a focus on the rear plate vs. the simulated picture.

### D. Comparative Experiments

Over these years, cameras' optical models have been used in many applications, including SFF/SFDF algorithms [20] and depth estimation [23]. However, these works usually use the parameters given in the hardware manual to generate the cameras' optical models, which leads to inaccurate models.

We compare our optical model with the optical model generated by the default parameters. For our camera (mentioned in Sect. 4.1), $F$ is 50, $\rho$ is 3.45 $\mu m$, $s$ is 1. We set the f-number ($N$) to 1.4, which is the maximum value of our lens. We set the $\omega$ to 0.48 [31]. We compute the optical parameter following (6) and its value is 6959. We use the optical parameter calculated by default parameters to generate sets of images, as shown in Fig. 11.

### D. Results and Discussion

The experimental results are shown in Fig. 9, Fig. 11 and Table 3. Intuitively, there are many differences between the images generated with the default parameters and the actual images, as shown in Fig. 11.

In Table 3, the $A$ and $e$ of our method are solved with the processes in Sect. 5. The $A$ of other method is calculated with the default value as described in Sect. 6.4. $e$ is set to the same value as ours, which is 23.6 mm. Seen from Table 3, the $A$ of our method is around 800, while the $A$ of the other method is 6959. There is a big difference between our method and others in $A$.

The big difference in optical parameter $A$ of two methods leads to quite different results. The simulated image in Fig. 11(c) and Fig. 11(f) has a much smaller depth of field. The transition from focus to defocus is more drastic than the actual images and the simulated images generated by our algorithm. This means the optical parameter calculated by the default parameters is not accurate enough. To qualitatively analyze the differences of the focused images, we compute the loss of actual focused images with the simulated images generated by our algorithm and others. As shown in Fig. 11 and Table. 3, the average loss of our method is 1336.17, while the average loss of the other method is 2217.97. The average loss between the actual focused images and the simulated images generated by our algorithm is 40% less than the loss between the actual images and the simulated images generated by others.

For the four losses we used, we can see that the loss3 plays the most important role. The average loss3 of our method is 0.013, but the average loss3 of others is 0.086. The latter is almost seven times larger than the former, which means the loss3 dominates the total loss. The results in all prove that our algorithm is closer to the actual imaging process.

TABLE 3

EXPERIMENTS RESULTS

| | Our method | | | | | | | Other method[1] | | | | | | |
| --- | --- | --- | --- | --- | --- | --- | --- | --- | --- | --- | --- | --- | --- | --- |
| | A | e | Loss | Loss1 | Loss2 | Loss3 | Loss4 | A | e | Loss | Loss1 | Loss2 | Loss3 | Loss4 |
| Mask1[2] | 800 | 23.6 | 1657.29 | 0.058 | 0.176 | 0.011 | 0.304 | 6959 | 23.6 | 2460.78 | 0.051 | 0.237 | 0.083 | 0.325 |
| Mask2 | 800 | 23.6 | 1552.28 | 0.058 | 0.167 | 0.012 | 0.232 | 6959 | 23.6 | 2417.33 | 0.055 | 0.232 | 0.079 | 0.284 |
| Mask3 | 800 | 23.6 | 1330.59 | 0.044 | 0.163 | 0.01 | 0.221 | 6959 | 23.6 | 2206.68 | 0.041 | 0.232 | 0.076 | 0.279 |
| Mask4 | 800 | 23.6 | 1510.46 | 0.052 | 0.168 | 0.009 | 0.279 | 6959 | 23.6 | 2302.18 | 0.047 | 0.229 | 0.075 | 0.313 |
| Object1[3] | 780 | 23.6 | 1103.37 | 0.028 | 0.148 | 0.012 | 0.212 | 6959 | 23.6 | 2070.11 | 0.028 | 0.189 | 0.094 | 0.244 |
| Object2 | 720 | 23.6 | 1305.30 | 0.026 | 0.140 | 0.017 | 0.369 | 6959 | 23.6 | 2214.49 | 0.022 | 0.175 | 0.105 | 0.361 |
| Object3 | 760 | 23.6 | 766.16 | 0.016 | 0.109 | 0.013 | 0.132 | 6959 | 23.6 | 1914.92 | 0.023 | 0.164 | 0.095 | 0.218 |
| Object4 | 800 | 23.6 | 1463.92 | 0.037 | 0.160 | 0.013 | 0.381 | 6959 | 23.6 | 2157.30 | 0.032 | 0.198 | 0.081 | 0.367 |
| **Avg** | | | 1336.17 | 0.040 | 0.154 | 0.012 | 0.299 | | | 2217.97 | 0.037 | 0.207 | 0.086 | 0.299 |

[1]It represents the model using the paraments given in the hardware manual, and using the same mechanical parameter.

[2]The standard plane with a simulated mask.

[3]The objects we used as shown in Fig. 9.

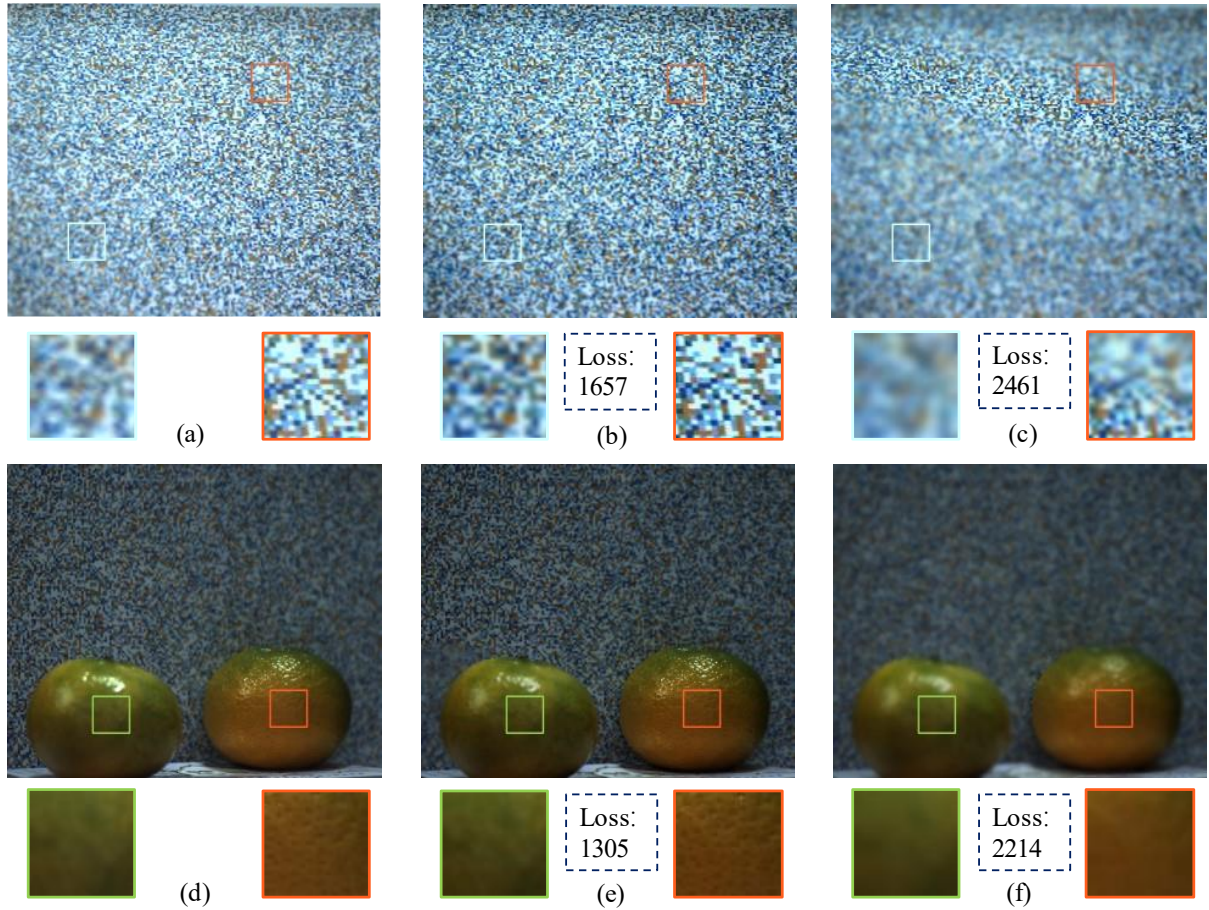

Fig. 11. The comparison of actual and simulated focused images. (a) and (d) are the actual focused images, (b) and (e) are the focused images generated by our method, (c) and (f) are the focused images generated by other method with default parameters.

## VII. CONCLUSIONS

The point spread function plays a crucial role in many fields. However, due to the lack of a dataset and a mathematical model, it is still a challenge of obtaining a precise PSF for a camera. We propose a precise dataset about all-in-focus images, focused images with corresponding focus depth, and depth images in the same view. On this basis, we derive a precise PSF for the camera.

We design a hardware system to obtain the dataset, which is hard to get before. Based on the dataset, we use a novel metric to evaluate the difference between the focused images to obtain a precise PSF. After obtaining the precise PSF, we compare the focused images generated by our algorithm with the actual focused images and the focused image generated by the previous algorithm, as shown in Fig. 11 and Table 3, to verify the accuracy of our model.

In all, we get an accurate dataset and a precise PSF model. We disclose the dataset in our experiments and keep them updated. We hope it will promote the development of related works, such as SFF/SFDF, image deconvolution, and depth estimation.

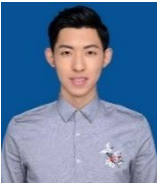 **Renzhi He** received the BS degree in mechatronics engineering form the Chongqing University of Posts and Telecommunications. He is currently studying in Chongqing University for his master's degree. His research interests include computational photography, computer vision, SFF, and structured light. He is a student member of the IEEE.

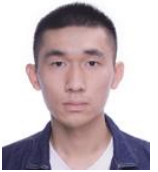 **Yan Zhuang** is currently studying in Chongqing University for his bachelor's degree in Mechanical Engineering. He is also studying in University of Cincinnati for his second bachelor's degree in Mechanical Engineering. He has worked in both R&D and IT department at Siemens. His research interests include deep neural networks and reinforcement learning.

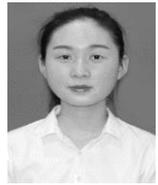 **Boya Fu** received the B.E. degree in machine design manufacture and automation from Southwest Jiaotong University, Chengdu, China, in 2020. She is currently studying as a graduate student in mechanical and electronic engineering in Chongqing University, Chongqing, China.
Her research interests are in computer vision, shape from focus and image processing.

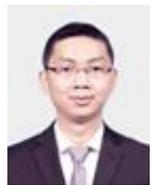 **Fei Liu** received his B.S. degree from Huazhong University of Science and Technology and Ph. D. degree from Tsinghua University, China, respectively. During 2012~2013, he was a visiting scholar in Massachusetts Institute of Technology, USA.
He is an associate professor with the Department of Mechanical Engineering at Chongqing University, China, and a Research Fellow in the State Key Lab of Mechanical Transmission. His research interests include robotics, machine vision, the optical measurement and its application.